\documentclass[letterpaper]{article} 
\usepackage{aaai25}  
\usepackage{times}  
\usepackage{helvet}  
\usepackage{courier}  
\usepackage[hyphens]{url}  
\usepackage{graphicx} 
\usepackage{natbib}  
\usepackage{caption} 
\usepackage{algorithm}
\usepackage{algorithmic}

\usepackage{newfloat}
\usepackage{listings}
\DeclareCaptionStyle{ruled}{labelfont=normalfont,labelsep=colon,strut=off} 
\floatstyle{ruled}
\newfloat{listing}{tb}{lst}{}
\floatname{listing}{Listing}
\usepackage{algorithm}
\usepackage{algorithmic}

\usepackage{graphicx}
\usepackage{booktabs}
\usepackage{multicol}
\usepackage{multirow}
\usepackage{amsmath}
\usepackage{tabularx}
\usepackage{subcaption}
\usepackage{cleveref}

\usepackage{newfloat}
\usepackage{listings}
\DeclareCaptionStyle{ruled}{labelfont=normalfont,labelsep=colon,strut=off} 
\floatstyle{ruled}
\newfloat{listing}{tb}{lst}{}
\floatname{listing}{Listing}

\title{PlanLLM: Video Procedure Planning with Refinable Large Language Models}
\author{
    Dejie Yang\textsuperscript{\rm 1},
    Zijing Zhao\textsuperscript{\rm 1},
    Yang Liu\textsuperscript{\rm 1,2}\footnote{Corresponding Author}
}
\affiliations{
    \textsuperscript{\rm 1} Wangxuan Institute of  Computer Technology, Peking University\\
    \textsuperscript{\rm 2} State Key Laboratory of General Artificial Intelligence, Peking University\\
    
    \{ydj,zijingzhao\}@stu.pku.edu.cn, yangliu@pku.edu.cn
}

\usepackage{xcolor}

\begin{document}

\maketitle
\begin{abstract}
Video procedure planning, \textit{i.e.}, planning a sequence of action steps given the video frames of start and goal states, is an essential ability for embodied AI. Recent works utilize Large Language Models (LLMs) to generate enriched action step description texts to guide action step decoding. Although LLMs are introduced, these methods decode the action steps into a closed-set of one-hot vectors, limiting the model's capability of generalizing to new steps or tasks. Additionally, fixed action step descriptions based on world-level commonsense may contain noise in specific instances of visual states. In this paper, we propose PlanLLM, a cross-modal joint learning framework with LLMs for video procedure planning. We propose an LLM-Enhanced Planning module which fully uses the generalization ability of LLMs to produce free-form planning output and to enhance action step decoding. We also propose Mutual Information Maximization module to connect world-level commonsense of step descriptions and sample-specific information of visual states, enabling LLMs to employ the reasoning ability to generate step sequences. With the assistance of LLMs, our method can  both closed-set and open vocabulary procedure planning tasks. Our PlanLLM achieves superior performance on three benchmarks, demonstrating the effectiveness of our designs.
Codes are available at: \url{https://github.com/idejie/PlanLLM}
\end{abstract}

\section{Introduction}
\label{sec:intro}

\begin{figure}[ht]
  \centering
  \begin{subfigure}{0.9\linewidth}
  \centering
    \includegraphics[width=\linewidth]{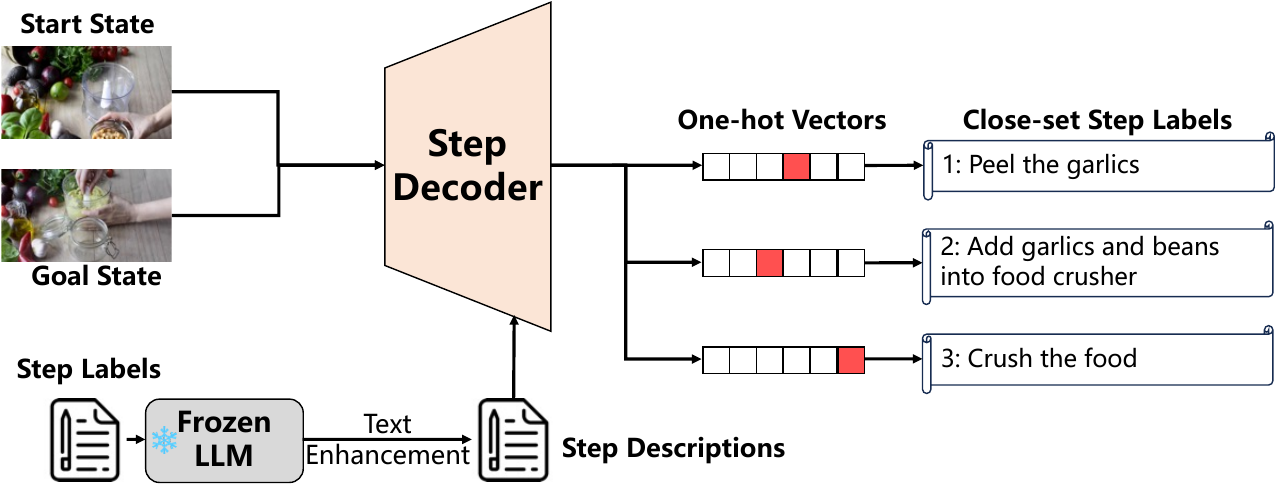}
    \setlength{\abovecaptionskip}{-0.3cm}
    \caption{Language  As Supervision}
    \label{fig:schema}
  \end{subfigure}
  \\
  \centering
  \begin{subfigure}{\linewidth}
  \centering
    \includegraphics[width=0.95\linewidth]{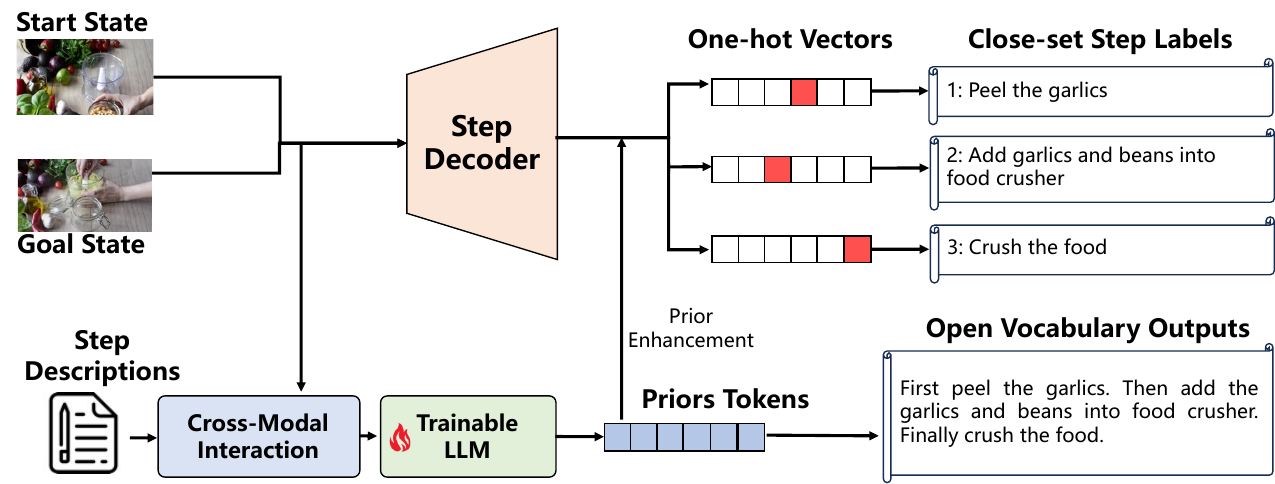}
    \setlength{\abovecaptionskip}{0.1cm}
    \caption{Cross-Modal Joint Learning (Ours)}
    \label{fig:ours}
  \end{subfigure}
  \setlength{\abovecaptionskip}{-0.2cm}
 \setlength{\belowcaptionskip}{-0.5cm}
  \caption{Language As Supervision methods use a frozen LLM to enhance the textual description of step labels, and are limited to closed-set one-hot vector predictions. In contrast, our Cross-Modal Joint Learning framework utilizes a \textbf{trainable LLM} that allows our model to output both one-hot vectors and \textbf{free-form open-vocabulary} step descriptions, providing stronger generalization for predicting new steps in new datasets. 
  }
  \label{fig:teaser}
\end{figure}

Mastering procedure planning is of great importance for building artificial intelligence systems capable of comprehending and imitating human actions, and eventually assisting humans in solving goal-directed problems  \cite{tellex2011understanding,jansen2020visually,ahn2022can,mishra2018tracking,mysore2019materials}. 
Previous research \cite{chang2020procedure} has pointed out that instructional videos are natural resources {for acquiring the skill} of procedure planning, proposing the problem of video procedure planning. 
For example, given the visual frames of start state ``\texttt{raw food}'' and goal states ``\texttt{hummus}'', the model is expected to produce a sequence of action steps to achieve the goal, i.e., first ``\texttt{peel the garlics}'', then ``\texttt{add garlics and beans into food crusher}'', and finally ``\texttt{crush the food}''.

{Early works employ both visual keyframes and textual labels of action steps to train the video procedure planning model (fully-supervised), which requires labor-intensive annotations.}
{Recent weakly-supervised methods rely only on the textual step sequences during training, and have gained more attention.}

Existing weakly supervised video procedure planning methods \cite{chang2020procedure,wang2023pdpp,li2023skip} use an encoder-decoder structure, first learning a latent visual space and then decoding action step labels from this space. 
These methods primarily treat video procedure planning as a visual-based action classification task on a closed-set of labels.
Recent approaches \cite{niu2023schema,zhao2022p3iv,wang2023event} introduce language models to {take advantage of} semantic information from textual action steps.
For instance, as shown in \Cref{fig:schema}, language as supervision method\cite{niu2023schema} uses a frozen large language model (LLM) to generate enriched descriptions of action steps based on world-level commonsense extracted from the LLM. By {combining} these language descriptions with visual states, the model is better trained to decode action steps, resulting in improved performance.

Though making progress, existing methods using LLMs still face issues:
(1) They use a step decoder that treats each action step as a one-hot vector in a closed set, which can not generalize to open-world scenarios.
Moreover, these approaches {ignore the} semantic relationships between steps. 
For example, ``\texttt{peel the food}'' and ``\texttt{crush the food}'' are likely to occur together, but decoding within a closed set cannot handle new steps or novel tasks.
(2) Methods that employ a frozen LLM generate consistent action step descriptions based on world-level commonsense, but these may not always be accurate for specific visual start and goal states. 
For instance, ``\texttt{crush the food}'' might generally suggest ``\texttt{mixed food pieces},'' but if the visual end state shows hummus in a food crusher, the correct outcome would be ``\texttt{hummus in the food crusher}.'' 
The lack of sample-specific visual cues limits the LLM's ability to apply contextual commonsense in step sequence reasoning.

To address the above challenges, we propose PlanLLM, a cross-modal joint learning framework with LLMs for video procedure planning. As shown in \Cref{fig:ours}, we utilize LLM's planning priors to enhance the model's generalization ability through an LLM Enhanced Planning module, producing free-form and open-vocabulary procedure planning outputs. Notably, we are the first to use a \textbf{trainable LLM} to output action step sequences that are \textbf{not confined to a predefined set} and can generalize to new steps or novel tasks.
To build cross-modal interaction between visual states and step descriptions, we introduce a Mutual Information Maximization {mechanism} to connect sample-specific visual information with world-level commonsense descriptions. This allows the LLM to use its generalizable reasoning ability, grounded in sample-specific visual commonsense, to enhance procedure learning.
We propose a progressive alignment scheme during training. 
{We first} freeze the LLM to align the visual embeddings to textual space using a mutual information maximization loss, {and then} fine-tune the LLM jointly with other modules for procedure planning.
Finally, the output from the LLM branch can contribute to conventional decoding by providing insights into {semantic relevance of steps, thereby} enabling our approach to handle both closed-set and open-vocabulary procedure planning tasks.

Overall, the {main} contributions of our work are as follows:
(1) We propose PlanLLM, a cross-modal joint learning framework with a trainable LLM for video procedure planning, which is the first to consider both the model's generalization ability in the open world and its planning performance on closed sets.
(2) We introduce the LLM-Enhanced Planning module, which utilizes the LLM's textual planning priors to enhance the capability of video procedure planning models. Additionally, we propose a Mutual Information Maximization module to connect world-level commonsense in step descriptions with sample-specific visual state information, enabling LLMs to utilize reasoning abilities for generating step sequences.
(3) PlanLLM achieves superior performance on three commonly used datasets: COIN \cite{tang2019coin}, CrossTask \cite{zhukov2019cross}, and NIV \cite{alayrac2016unsupervised}, and improves cross-dataset procedure planning performance, demonstrating the effectiveness and generalization of our method.

\section{Related Works}
\label{sec:relate}

\paragraph{Video Procedure Planing.}
{Video procedure planning task \cite{chang2020procedure}} aims to generate a sequence of action steps based on visual observations of the start and goal states. Early works~\cite{srinivas2018universal,chang2020procedure,bi2021procedure,sun2022plate} train the models in a fully supervised manner, requiring annotations for textual step sequences and intermediate visual states. Recently, weakly supervised approaches, which rely only on textual step sequences during training, have gained attention due to their lower annotation costs.
These methods \cite{chang2020procedure,wang2023pdpp,li2023skip,sun2022plate} typically use an encoder-decoder structure to learn a latent visual space and decode action step labels from it. 
PDPP \cite{wang2023pdpp} {introduces} a diffusion probabilistic model for generating intermediate action labels, while KEPP \cite{Nagasinghe_2024_CVPR} {proposes} a probabilistic procedural knowledge graph to improve planning. 
Instead of focusing on visual representations, several methods \cite{zhao2022p3iv,wang2023event,liu2023language} utilize language encoders to capture the textual information of action steps.
SCHEMA \cite{niu2023schema} leverages large language models like GPT-3.5 to extend the descriptions of textual action step labels, {then uses the extended descriptions} to guide the decoding of action steps.

In this paper, we achieve {weakly supervised video procedure planning} relying only on textual action step labels during training. 
Unlike existing methods that decode action steps into a closed-set of one-hot vectors, we harness the generalization ability of LLMs to generate free-form procedure planning outputs. 
Our approach not only enhances the decoding of {closed-set} labels, but also allows the model to generate free-form outputs for new action steps and planning tasks.
Note that the step descriptions used in both our method and the Language As Supervision approaches {such as \cite{niu2023schema}} are dataset-specific, aggregated from step labels and enhanced by the LLM, and are independent of the sample-specific visual states.

\begin{figure*}
  \centering
  \includegraphics[width=0.95\linewidth]{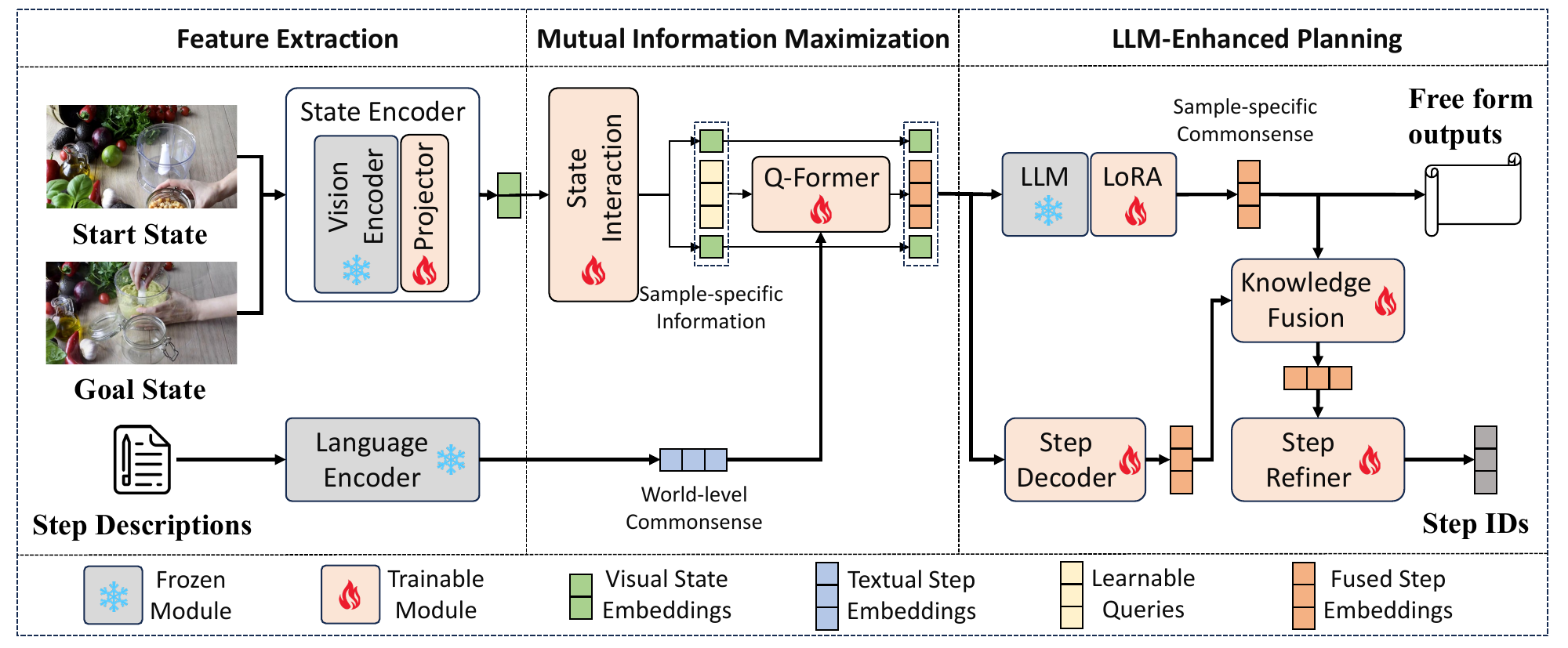}
    \setlength{\abovecaptionskip}{-0.0cm}
     \setlength{\abovecaptionskip}{0.1cm}
 \setlength{\belowcaptionskip}{-0.5cm}
\caption{\textbf{The framework of our PlanLLM.} PlanLLM mainly consists of three parts: Feature Extraction, Mutual Information Maximization and LLM Enhanced Planning. 
 }
\label{frame}
\end{figure*}

\paragraph{LLMs for Planning.}
Pre-trained large language models (LLMs) \cite{brown2020language,chowdhery2023palm} have demonstrated their ability to generate high-level  plans~\cite{xie2023translating,wang2023describe,liu2023llava,ye2024cat,Yang_2024_CVPR} and video understanding\cite{SPL_2023_ACL} . 
Some works train LLMs to generate verbal reasoning traces and text actions\cite{yao2022react,Lei_2024_CVPR} in an interleaved manner, allowing the model to interface with external sources (e.g., knowledge bases or environments) for additional information. 
{Other works  leverages LLMs for few-shot planning  \cite{song2023llm}, interaction~\cite{ting2024CMMP}  and temporal understanding~\cite{zheng2024TFTG}, enhancing them with physical grounding to generate and update plans based on the current environment. 
These enhanced LLMs can follow natural language instructions to complete complex tasks in visually-perceived environments.
{Although achieving success}, these methods treat LLMs as textual planners, relying solely on textual descriptions of the environment, which may not fully capture the complexity of visual states, thus limiting visual reasoning abilities.

Unlike existing methods that focus on textual planning with LLMs, our approach exploits the capabilities of LLMs for video procedure planning. 
We connect sample-specific visual state information with world-level commonsense from LLMs, training the model to learn multi-modal representations for a better understanding of the environment.

\section{Methodology}
\label{sec:method}

\subsection{Problem Formulation and Overview}
Video procedure planning task aims to generate a sequence of action steps based on visual observations of the start and goal states. 
As shown in Figure \ref{frame}, given the visual observations (images) of the start frame $v_0$ and the goal $v_T$, the target is to plan a procedure, represented as a sequence of action steps $\hat{\pi} = a_{1:T}$, which are sequentially performed to achieve the transition from $v_0$ to $v_T$. 
The video procedure planning problem can be formulated as $p(a_{1:T}|v_0, v_T)$.
We follow the weakly supervised video procedure planning setting \cite{wang2023pdpp,niu2023schema}, which rely only on the ground truth textual action sequence $a_{1:T}$ without requiring corresponding intermediate visual states (keyframes between two states). 
{Following previous work \cite{niu2023schema}}, we use the textual information of action step labels, treating video procedure planning as a multi-modal task.
Notably, the textual information of step descriptions are collected from the training set and shared across different visual state inputs, making them independent of specific inputs.

As depicted in \Cref{frame}, our PlanLLM encompasses three parts:
(1) \textit{Feature Extraction}: We use a visual encoder and a language encoder to extract features from the visual inputs $\{v_0, v_T\}$ representing the start and goal states, capturing sample-specific {visual details essential for task context}, and from the textual descriptions $D$ of action steps, capturing world-level commonsense.
(2) \textit{Mutual Information Maximization}: To {fuse} sample-specific visual information with world-level textual commonsense, we introduce {Q-Former architecture which takes} visual embeddings processed by the state interaction module and step description embeddings as inputs to generate cross-modal joint step embeddings.
(3) \textit{LLM Enhanced Planning}: To address the limitations of one-hot decoding of action steps, which ignores semantic relevance between steps and cannot generalize to new steps or tasks, we propose the LLM-Enhanced Planning module. This module takes multi-modal step embeddings from the Mutual Information Maximization module as inputs, allowing the LLM to be fine-tuned via LoRA \cite{hu2022lora} to directly produce free-form procedure planning outputs, capable of handling new steps and tasks. 
Additionally, {when} combined with our proposed knowledge fusion and step refinement module, the general reasoning ability of LLMs {is effectively leveraged to refine} the original step decoding process, improving performance.

\subsection{Feature Extraction}
We first introduce a visual state encoder $E_v$ and a language encoder $E_l$ to extract features from sample-wise visual state inputs $\{v_0, v_T\}$ and world-level action step descriptions $D$, respectively.

\textit{Visual State Feature Extraction:}
The visual state encoder takes the video frames of the start and goal states as input and outputs embeddings that represent sample-specific visual information. We adopt a pre-trained vision backbone $f_v$ together with a trainable projection layer $p_v$ as the visual state encoder $E_v$. The embedding $x_t$ of visual state $v_t, t \in \{0, T\}$ is extracted by:
\begin{align}
    x_t &= E_v{(v_{t})}=p_v(f_v(v_{t})), ~t \in \{0, T\}
\end{align}

\textit{Textual Step Feature Extraction:}
As the original step texts are too concise, following \cite{niu2023schema}, we extend all the action step labels $A=\{a_i\}_{i = 1}^{N}$ to enriched action step descriptions $D=\{d_i\}_{i=1}^N$ using large language models (such as GPT-3.5) with world-level commonsense knowledge:
\begin{align}
    d_i = \mathrm{LLM}(prompt, a_i), ~ i \in \{1, 2, ..., N\}
\end{align}
where $N$ denotes the number of all possible action step in the dataset.
The details of description generation are provided in supplemental materials.
We adopt a fixed pre-trained language encoder $E_l$ to extract the embedding $y_i$ of the action step descriptions $d_i$:
\begin{equation}
    y_i = E_l(d_i)
    \label{y_i}
\end{equation}

\subsection{Mutual Information Maximization}
As the visual state embeddings contain only sample-specific information while the textual step embeddings capture only world-level commonsense knowledge from LLMs, we propose a Mutual Information Maximization module to integrate both embeddings for a better representation of action steps.

\textit{Visual State Interaction:}
To learn the mutual context awareness of visual states, we introduce a self-attention layer to process the original embeddings $x_0^v, x_T^v$ to interacted embeddings $x_0^{v}, x_T^{v}$:
\begin{equation}
    x_0^{v}, x_T^{v} = \mathrm{SelfAtten}\left(x_0, x_T\right)
\end{equation}
Interacted visual embeddings provide a better representation when integrated with world-level textual step embeddings and when aligned with LLM input space.

\textit{{Mutual Information Maximization}(MIM) Q-Former:}
To integrate sample-specific visual state embeddings and textual step embeddings with world-level commonsense knowledge, we adopt a Q-Former \cite{li2023blip} architecture to learn fused step embeddings. Specifically, we use the integrated visual start and goal state embeddings $x_0^{v}, x_T^{v}$ as the image input tokens for Q-Former, and the textual action step embeddings $x_{1:T}^a$ corresponding to the ground-truth step sequence $a_{1:T}$ as the textual input tokens. Using learnable step queries $q_{1:T}$, the Q-Former outputs the integrated step embeddings $x_{1:T}^{q}$:
\begin{equation}
    x_{1:T}^q = \mathrm{QFormer}\left(q_{1:T};~[x_0^{v}, x_T^{v}];~a_{1:T}\right)
\end{equation}
Following Q-Former, we optimize the Vision-Language Contrastive (VLC) loss and Vision-Language Matching (VLM) loss to maximize the mutual information between vision and language embeddings. Within a batch, we treat the visual state embeddings and their corresponding ground truth step embeddings as positive pairs, and unmatched vision and textual embeddings as negative pairs. The Vision-Language Contrastive (VLC) loss contrasts the embedding similarity of a positive pair against the negative ones:
\begin{equation}
    \mathcal{L}_{VLC} = - \log  \frac{\sum_j e^{{s\left(x^{v}[j], x^{q}[j]\right)}}}{\sum_j e^{{s\left(x^{v}[j], x^{q}[j]\right)}} + \sum_{j\neq k} e^{{s\left(x^{v}[j], x^{q}[k]\right)}}}
\end{equation}
where $s(\cdot, \cdot)$ denotes the similarity function, and $x^{v}[j]$ and $x^q[j]$ represent the interacted visual state embeddings from the visual state interaction module and the integrated step embeddings from Q-Former, respectively, for the $j$th sample in a batch, with $j,k \in \{1, 2, \dots, B\}$, where $B$ is the batch size. 
For the Vision-Language Matching (VLM) loss, the model is tasked with a binary classification to predict whether an image-text pair is positive (matched) or negative (unmatched):

\begin{equation}
    \mathcal{L}_{VLM} = \sum_{j,k} \mathcal{L}_{BCE}\left(x^{v}[j], x^q[k]; \mathrm{\mathbf{1}}(j=k)\right)
\end{equation}
The total Mutual Information Maximization (MIM) loss combines VLC loss and VLM loss:
\begin{equation}
    \mathcal{L}_{MIM} = \mathcal{L}_{VLC} + \mathcal{L}_{VLM}.
    \label{loss_mim}
\end{equation}

\subsection{LLM Generated Free Form Planning}
Different from previous methods that decode action steps into a closed set of one-hot vectors, our approach fully leverages the LLM's generalization ability to directly generate free-form procedure planning outputs, enabling it to handle open vocabulary procedure planning and new planning tasks.

\textit{Free Form Procedure Planning Output:}
We introduce a generative LLM, \textit{i.e.}, Vicuna-7B \cite{vicuna2023}. Taking the integrated step embeddings $x_{1:T}^q$ from the MIM Q-Former along with the visual state embeddings $x_0^{v}, x_T^{v}$ as inputs, the LLM's encoder provides hidden state embeddings $h_{1:T}$, representing the enhanced action step embeddings with sample-specific commonsense, based on the general reasoning ability of the LLM. Its decoder then directly generates free-form output tokens $O$ as the captioning of the action steps:

\begin{align}
    h_{1:T} &= \mathrm{LLM}_{enc}\left(x_0^{v}, x_T^{v}, x_{1:T}^q\right) \\
    O &= \mathrm{LLM}_{dec}\left(h_{1:T}\right)
\end{align}

\textit{Progressive Alignment Training Scheme}:
To effectively train the modules to produce free-form procedure planning outputs, we propose a two-stage progressive training scheme. Specifically, in the first stage, we fix the generative LLM and train the feature extractors and MIM modules using $\mathcal{L}_{MIM}$ as defined in Equation \ref{loss_mim}. This stage aligns the visual state embeddings and fused action step embeddings with the LLM's input space, enabling the LLM to understand and reason with this information.
In the second stage, we fine-tune the LLM using LoRA \cite{hu2022lora}, along with other trainable modules, optimizing an Action Step Captioning loss $\mathcal{L}_{ASC}$ and the original MIM loss $\mathcal{L}_{MIM}$. The $\mathcal{L}_{ASC}$ loss enables the LLM to learn step caption generation, with the ground truth action step texts organized into a formatted sentence.
With our progressive alignment training scheme, the LLM is effectively trained to produce free-form procedure planning outputs.

\subsection{LLM Enhanced Close-set Step Decoding}
{Existing methods decode action steps as a closed set of one-hot vectors.} With the help of LLMs which understand the semantic information of different action step labels, the conventional action step decoding process can also be enhanced to improve performance.

\textit{Action Step Decoder:}
{Following previous work \cite{niu2023schema}, we employ a cross-attention module to process} the visual state embeddings, action step description embeddings, and learnable queries to produce final action step representations, which are then passed through a classifier to distinguish specific action steps. 
The final action step representations $r_{1:T}^{SD}$ are obtained by:
\begin{align}
    r_{1:T}^{SD} &= \mathrm{CrossAtten}\left( q_{1:T}^{SD}, [x_0^{v}, x_T^{v}, x_{1:T}^q]\right) 
\end{align}
where $x_0^{v}, x_T^{v}$ denotes the visual start and end state embeddings, $x_{1:T}^q$ denotes fused action step embeddings from MIM Q-Former, and $q_{1:T}^{SD}$ denotes learnable queries.

\textit{Knowledge Fusion:}
To integrate the semantic understanding knowledge of LLMs, we fuse the sample-specific commonsense of hidden states embeddings $h_{1:T}$ from the LLM encoder with the final action step representations $r_{1:T}^{SD}$ from the step decoder to produce step representations:
\begin{align}
    r_{1:T}^{KF} &= \mathrm{CrossAtten}\left(r_{1:T}^{SD}; h_{1:T}\right) 
\end{align}
With knowledge fusion module, the knowledge fused step representations are aware of the semantic relevance between action step labels with the help of LLM general knowledge.

\textit{Step Refiner:}
Given the knowledge fused representations $r_{1:T}^{KF}$ as query, we finally refine the step representations from the embeddings of the action step descriptions with another cross-attention module :
\begin{align}
    r_{1:T}^{SR} = \mathrm{CrossAtten}\left(r_{1:T}^{KF}; y\right), 
\end{align}
where $y$ is the embeddings of the action step descriptions via \Cref{y_i}.
The refined step representations $r_{1:T}^{SR}$ are projected into a one-hot vector $\hat{a}_{1:T}$ for classification, supervised by cross-entropy loss as the Step Decoding (SD) loss:
\begin{align}
    \mathcal{L}_{SD} = \mathcal{L}_{CE}(\hat{a}_{1:T}, a_{1:T})
\end{align}
Where $\mathcal{L}_{CE}$ denotes cross-entropy loss, and $a_{1:T}$ denotes ground truth action step labels.

\subsection{Training and Inference}
\textit{Training:}
As has been discussed in \textit{Progressive Alignment Training Scheme}, in the first stage, we fix the visual backbone encoder, the language encoder and the generative LLM, and only train the Mutual Information Maximization Module with $\mathcal{L}_{MIM}$ in Equation \ref{loss_mim} to align the visual state embeddings and the action step embeddings to the LLM input space.
In the second stage, we jointly fine-tune the LLM through LoRA\cite{hu2022lora} and the other modules with the corresponding losses.
{The overall objective can be elaborated as}:
\begin{equation}
    \mathcal{L} = \mathcal{L}_{MIM} +  \mathcal{L}_{ASC} + \mathcal{L}_{SD}
\end{equation}
where $\mathcal{L}_{ASC}$ is the Action Step Captioning loss, $\mathcal{L}_{SD}$ is the State Decoding loss.

\textit{Inference:}
Our method can handle both closed-set action step classification tasks and free-form open-vocabulary procedure planning tasks. For conventional action step classification, our proposed LLM-Enhanced step decoding branch outputs the action step IDs following \cite{niu2023schema,wang2023pdpp,zhao2022p3iv}. 
For open-vocabulary procedure planning, the generative LLM provides free-form procedure planning outputs and encodes the captions and new textual action step labels into vectors using a frozen language encoder. We then retrieve the top $T$ action labels based on the similarity between captions and textual action step labels, where $T$ is the number of action steps in a sequence. 

\begin{table*}[t]
  \centering
  \resizebox{0.8\linewidth}{!}{
  \begin{tabular}{cc|ccc|ccc}
    \toprule
   \multirow{2}{*}{Method} &\multirow{2}{*}{Supervision } & \multicolumn{3}{c|}{t=3} &\multicolumn{3}{c}{t=4}\\
   & &SR$\uparrow$&mAcc$\uparrow$&mIoU$\uparrow$&SR$\uparrow$&mAcc$\uparrow$&mIoU$\uparrow$\\
    \midrule
   
\multicolumn{1}{l}{UPN\cite{srinivas2018universal}}              &      V+A       &   2.89    &    24.39    &   {31.56}     &   1.19    &   21.59     &   {27.85}\\
\multicolumn{1}{l}{DDN\cite{chang2020procedure}}              &      V+A     &   12.18    &    31.29    &    {47.48}    &   5.97    &    27.10    &    {48.46}  \\
\multicolumn{1}{l}{Ext-GAIL\cite{bi2021procedure}}         &      V+A       &   21.27    &   49.46     &   {61.70}     &   16.41    &    43.05    &   {60.93}   \\
\multicolumn{1}{l}{PlaTe\cite{sun2022plate}}&V+A& 16.00 &36.17&{65.91} &14.00 &35.29&{55.36}\\
    \midrule
\multicolumn{1}{l}{P$^3$IV\cite{zhao2022p3iv}}             &      A      &   23.34    &   49.96     &   {73.89}     &   13.40    &   44.16     &   {70.01}  \\
 \multicolumn{1}{l}{EGPP \cite{wang2023event}}           & A    & 20.14  & 38.36  & {67.29} & 11.32  & 18.85  & {70.53}\\
\multicolumn{1}{l}{PDPP\cite{wang2023pdpp}}             &      A       &    {37.20}   &   {64.67}    &     {66.57}   &   {21.48}    &   {57.82}     &    {65.13} \\
\multicolumn{1}{l}{SkipPlan\cite{li2023skip}}             &      A       &    {28.85}   &    {61.18}    &     {74.98}   &   {15.56}    &   {55.64}     &    {70.30}\\
\multicolumn{1}{l}{LangFirst\cite{liu2023language}}             &      A       &    {25.01}   &    {53.79}    &     {75.43}   &   {14.11}    &   {47.93}     &    {73.21}\\
\multicolumn{1}{l}{KEPP\cite{Nagasinghe_2024_CVPR}}             &      A       &    {38.12}   &    \underline{64.74}    &     {{67.15}}   &  {24.15}    &   \bf{59.05}     &    {{66.04}}  \\
\multicolumn{1}{l}{SCHEMA\cite{niu2023schema}}             &      A       &    \underline{38.93}   &    {63.80}    &     \underline{{79.82}}   &  \underline{24.50}    &   {58.48}     &    \underline{{78.42}} \\       
\multicolumn{1}{l}{PlanLLM(Ours)}             &      A       &    \textbf{39.74}   &    \textbf{65.78}    &     \textbf{{81.50}}   &  \textbf{27.54}    &   \underline{59.01}     &    \textbf{{77.58}}  \\
  \bottomrule
  \end{tabular}}
   \setlength{\abovecaptionskip}{0.1cm}
                \setlength{\belowcaptionskip}{-0.3cm}
  \caption{
   Comparisons on CrossTask for procedure planning with prediction horizon $t \in \{3, 4 \}.$ \textit{Supervision} denotes the supervision type, where $V$ denotes the methods use intermediate visual states (frames between start and goal states) as  supervisions, and $A$ only uses the action or task category without visual states. }
  \label{tab:CrossTask}
\end{table*}

\begin{table*}[tb]

\centering
  \resizebox{0.86\linewidth}{!}{
\begin{tabular}{ccc|ccc|ccc}
\toprule
              \multirow{2}{*}{Horizon}            &  \multirow{2}{*}{Method} &\multirow{2}{*} {Supervision } & \multicolumn{3}{c|}{NIV} & \multicolumn{3}{c}{COIN} \\  
              
&&& SR$\uparrow$     & mAcc$\uparrow$   & mIoU$\uparrow$  & SR$\uparrow$     & mAcc$\uparrow$   & mIoU$\uparrow$   \\ \hline
\multirow{10}{*}{$t$ = 3} 
                     & \multicolumn{1}{l}{DDN \cite{chang2020procedure}}            & V+A    & 18.41  & 32.54  & {56.56} & 13.9   & 20.19  & {64.78}  \\
                     & \multicolumn{1}{l}{Ext-GAIL \cite{bi2021procedure}}       & V+A    & 22.11  & 42.20  & {65.93} & -      & -      & -      \\
                     \cmidrule {2 -9}
                     & \multicolumn{1}{l}{P$^3$IV \cite{zhao2022p3iv}}           & A    & 24.68  & \underline{49.01}  & {74.29} & 15.4   & 21.67  & {76.31}  \\
                     & \multicolumn{1}{l}{EGPP \cite{wang2023event}}           & A    & 26.05 & \textbf{51.24}  & {75.81} & 19.57  & 31.42 & {\underline{84.95}}  \\
                     & \multicolumn{1}{l}{PDPP\cite{wang2023pdpp}}            & A    & \bf{30.20}  & {48.45}  & {57.28} & {21.33}  & {45.62}  & {51.82}  \\ 
                     &\multicolumn{1}{l}{SkipPlan\cite{li2023skip}}             &      A       &    -  &    -   &    -   &   {23.65}    &   {47.12}     &    {78.44}    \\
                     &\multicolumn{1}{l}{LangFirst\cite{liu2023language}}             &      A       &    -  &   -   &    -   &   {28.35}    &    \underline{53.14}     &    {78.56}    \\
                    & \multicolumn{1}{l}{SCHEMA\cite{niu2023schema}}             &      A       &    \underline{27.93 }   &    {41.64}    &     {{76.77}}   &   \underline{32.09}    &  {49.84}     &    {83.83}    \\
                    &\multicolumn{1}{l}{KEPP\cite{Nagasinghe_2024_CVPR}}             &      A       &    {24.44}   &    {43.46}    &     \textbf{86.67}   &   {20.25}    &   {39.87}     &    {-}    \\
                    & \multicolumn{1}{l}{PlanLLM(Ours)}             &   A       &    {26.74}  &    {42.97}   &     \underline{{ 77.23}}   &   \textbf{33.22}    &   \textbf{54.33}    &    \textbf{85.21} \\
                    \midrule
\multirow{10}{*}{$t$ = 4} 
                     & \multicolumn{1}{l}{DDN \cite{chang2020procedure}}            & V+A   & 15.97  & 27.09  & {53.84} & 11.13  & 17.71  & {68.06}  \\
                     & \multicolumn{1}{l}{Ext-GAIL \cite{bi2021procedure}}       & V+A    & 19.91  & 36.31  & {53.84} & -      & -      & -      \\
                     \cmidrule {2 -9}
                     & \multicolumn{1}{l}{P$^3$IV \cite{zhao2022p3iv}}           & A    & 20.14  & 38.36  & {67.29} & 11.32  & 18.85  & {70.53}  \\
                      & \multicolumn{1}{l}{EGPP \cite{wang2023event}}           & A    & 21.37  & 41.96  & {74.90} & 13.59  & 26.72  & {\underline{84.72}}  \\
                     & \multicolumn{1}{l}{PDPP\cite{wang2023pdpp}}            & A    & \underline{26.67}  & \underline{46.89}  & {59.45} & {14.41}  & {44.10}  & {51.39}  \\
                     &\multicolumn{1}{l}{SkipPlan\cite{li2023skip}}             &      A       &    -  &    -   &    -   &   \underline{23.65}    &   \underline{47.12}     &    {78.44}    \\
                    & \multicolumn{1}{l}{LangFirst\cite{liu2023language}}             &      A     &    -  &   -   &    -     &    {16.04}   &    {43.19}    &     {77.07}   \\
                     &\multicolumn{1}{l}{SCHEMA\cite{niu2023schema}}             &      A       &    {23.26}   &    {39.93}    &     {76.75}   &   {22.02}    &   {45.33}     &    {83.47}    \\
                     &\multicolumn{1}{l}{KEPP\cite{Nagasinghe_2024_CVPR}}             &      A       &    {22.71}   &    {41.59}    &     \textbf{91.49}   &   {15.63}    &   {39.53}     &    {-}    \\
                     & \multicolumn{1}{l}{PlanLLM(Ours)}             &      A       & \textbf{27.08} & \bf{46.96}&    \underline{{ 77.89}}   &  \textbf{25.31}    &   \textbf{{48.79}}    &    \textbf{86.28} \\
                     \bottomrule
\end{tabular}}
 \setlength{\abovecaptionskip}{0.1cm}
                \setlength{\belowcaptionskip}{-0.3cm}
\caption{
Evaluation results on NIV and COIN with prediction horizon $t \in \{3,4\}$.\label{tab:coin_niv} 
} 
\end{table*}

\begin{table*}[t]

\centering
{
\begin{tabular}{l|cc|ccc|ccc}
\hline
               \multirow{2}{*}{Method} &Train &Test& \multicolumn{3}{c|}{t=3} & \multicolumn{3}{c}{t=4} \\  
              
&Data&Data& SR$\uparrow$     & mAcc$\uparrow$   & mIoU$\uparrow$  & SR$\uparrow$     & mAcc$\uparrow$   & mIoU$\uparrow$   \\ \hline

                          SCHEMA     &  COIN &CrossTask &5.37&17.21&54.52&4.45&11.32&50.26 \\
                          
             PlanLLM(Ours) &  COIN &CrossTask &12.13&19.79&58.32& 8.32&14.13&53.44 \\
             \midrule
                          SCHEMA     &  CrossTask &COIN &3.23 &10.56&49.21&1.45&9.63&46.51\\
                          
             PlanLLM(Ours) &  CrossTask &COIN  &9.97&12.43&51.17&7.18&12.21&52.21\\
 
\hline
                \end{tabular}}
                 \setlength{\abovecaptionskip}{0.1cm}
                \setlength{\belowcaptionskip}{-0.3cm}
\caption{Performance  comparisons on cross-dataset with prediction horizon $t \in \{3,4\}$. } \label{tab:cross}
                \end{table*}

\begin{table}[ht]

\centering
\renewcommand\arraystretch{1}
\resizebox{\columnwidth}{!}{
\begin{tabular}{l|ccc|ccc}
\hline
               \multirow{2}{*}{Method}  & \multicolumn{3}{c|}{t=3} & \multicolumn{3}{c}{t=4} \\  
              
& SR$\uparrow$     & mAcc$\uparrow$   & mIoU$\uparrow$  & SR$\uparrow$     & mAcc$\uparrow$   & mIoU$\uparrow$   \\ \hline
$w/o$ MIM/LLM    & 38.03& 62.71&79.10&21.86&56.28&74.81  \\
$w/o$ LLM & 38.99& 63.23&79.76&22.42&57.25&75.17\\
             $w/o$ MIM    & 39.03& 64.12&80.25&23.24&57.94&76.11  \\
             
                            Full   &    \textbf{39.74}   &    \textbf{65.78}    &     \textbf{{81.50}}    &  \textbf{27.54}    &   \textbf{59.01}     &    \textbf{{77.58}}    \\
\hline
                \end{tabular}}
                \setlength{\abovecaptionskip}{0.1cm}
                \setlength{\belowcaptionskip}{-0.3cm}
\caption{Effectiveness of proposed components}
\label{tab:component}
                \end{table}

 \begin{table}[ht]

\centering
\renewcommand\arraystretch{1}
\resizebox{\columnwidth}{!}{
\begin{tabular}{l|ccc|ccc}
\hline
               \multirow{2}{*}{Method}  & \multicolumn{3}{c|}{t=3} & \multicolumn{3}{c}{t=4} \\  
              
& SR$\uparrow$     & mAcc$\uparrow$   & mIoU$\uparrow$  & SR$\uparrow$     & mAcc$\uparrow$   & mIoU$\uparrow$   \\ \hline

             Frozen LLM & 37.68& 61.26&77.48&22.92&56.19&74.34\\
             One-Stage& 38.83&62.17&77.82&23.55&57.67&75.21\\
                           Progressive   &      \textbf{39.74}   &    \textbf{65.78}    &     \textbf{{81.50}}  &  \textbf{27.54}    &   \textbf{59.01}     &    \textbf{{77.58}}   \\ 
\hline
                \end{tabular}}
\setlength{\abovecaptionskip}{0.1cm}
\setlength{\belowcaptionskip}{-0.3cm}
\caption{Effectiveness of progressive multi-modal training }

\label{tab:progressive}
\end{table}

\begin{table}[ht]

\centering
\renewcommand\arraystretch{1}
\resizebox{\columnwidth}{!}{
\begin{tabular}{l|ccc|ccc}
\hline
               \multirow{2}{*}{Method}  & \multicolumn{3}{c|}{t=3} & \multicolumn{3}{c}{t=4} \\  
              
& SR$\uparrow$     & mAcc$\uparrow$   & mIoU$\uparrow$  & SR$\uparrow$     & mAcc$\uparrow$   & mIoU$\uparrow$   \\ \hline

             LLM only    &  37.76&63.25&79.81&24.76&58.01&76.13 \\
             Step  only & 38.21&63.52&79.13&24.03&58.17&75.98\\
                            Fusion   &    \textbf{39.74}   &    \textbf{65.78}    &     \textbf{{81.50}} & \textbf{27.54}    &   \textbf{59.01}     &    \textbf{{77.58}}     \\ 
\hline
                \end{tabular}}
 \setlength{\abovecaptionskip}{0.1cm}
 \setlength{\belowcaptionskip}{-0.3cm}
\caption{Different planning generation strategy}
\label{tab:generation}
                \end{table}

\section{Experiments}
\label{sec:exp}

\subsection{Evaluation}

\subsubsection{Datasets}
For our evaluation, we employ three {commonly used} instructional video datasets: CrossTask~\cite{zhukov2019cross}, NIV~\cite{alayrac2016unsupervised}, and COIN~\cite{tang2019coin}.
The CrossTask dataset comprises 2,750 videos, illustrating 18 unique procedural tasks. 
{NIV (Narrated Instructional Videos) contains} 150 videos encompassing five procedures. 
COIN stands out as the largest dataset in our evaluation, boasting 11,827 videos, covering 778 procedures.

\subsubsection{Metrics}
We evaluate performance using three  metrics  as outlined in \cite{niu2023schema}:
(1) Mean Intersection over Union (mIoU)  measures whether the model correctly identifies the set of steps required to execute the plan, regardless of the order of actions.
(2) Mean Accuracy (mAcc) compares the predicted and actual action sequences element-wise, considering the correct order of actions.
(3) {Success Rate (SR) assesses the plan's success only if it exactly matches the ground truth, requiring precise correspondence between the predicted and actual sequences.}
Note that SR is a {stricter} metric, offering a stringent assessment of the model's performance and requires exact correspondence between the predicted and actual sequences for a plan to be successful. 

\subsubsection{Implementation details}
Following recent advancements \cite{niu2023schema,zhao2022p3iv,wang2023pdpp,wang2023event}, we leverage a pretrained S3D network \cite{miech2020end} as the visual backbone and the textual encoder of the pretrained CLIP \cite{radford2021learning} as our language encoder. Additionally, we deploy the trainable Q-Former architecture initialized from BLIP2 \cite{li2023blip}. Following \cite{li2023mvbench}, we utilize the pretrained Vicuna\cite{vicuna2023} as the Large Language Model (LLM).
During the frozen-LLM training stage, we set the learning rate to $1\times 10^{-4}$ for the Q-Former and $1\times 10^{-3}$ for other modules, training the model with a batch size of 32 on NVIDIA A800 GPUs.

\subsection{Results}

\subsubsection{Comparisons on CrossTask}
\Cref{tab:CrossTask} shows the comparisons between our method and others on CrossTask.
(1) Compared to previous fully-supervised methods \cite{ehsani2018let,abu2019uncertainty} that require intermediate visual states as inputs, recent weakly-supervised methods \cite{zhao2022p3iv,li2023skip,liu2023language,niu2023schema} achieve better results due to more advanced model designs. We also follow the weakly-supervised setting in our method.
(2) {Methods \cite{zhao2022p3iv, li2023skip, liu2023language, niu2023schema} that use embeddings of action or task categories (Supervision is A) are better suited for video procedure planning than those using intermediate visual states as supervision (Supervision is V).} This may be because embeddings offer more robust language representations for planning, unlike visual states, which have limited quantity and representation.
(3) The performance of the models is better at $t=3$ than $4$ {since the longer action sequence increases prediction difficulty.}
(4) {At horizon $t=3$, we surpass the previous best by +0.81\% on SR (a stricter, order-sensitive metric).} At $t=4$, our method also shows gains on SR (+3.04\%). 
These results demonstrate the effectiveness of incorporating LLMs in video procedure planning.

\subsubsection{Comparisons on NIV and COIN}
\Cref{tab:coin_niv} shows the comparisons between our method and others on NIV and COIN datasets.
(1) Similar to the results on CrossTask, the  methods using the embeddings of {action or task categories} obtain better performance  thanks to the powerful representation ability of language encoders. 
(2) On NIV, compared to previous best performance, our method obtains gains of +0.41\% on SR(a more strict metric, order-sensitive) when prediction horizon $t=4$.
This may be because LLM can provide commonsense knowledge for long action sequences, {improving} the accuracy of procedure planning.
(3) On COIN dataset, our method remains the best performance across metrics. Specifically, our method makes improvement on SR (+1.13\%), mAcc(+1.19\%) and mIoU(+0.26\%), when prediction horizon $t=3$. And our PlanLLM also has gains (+1.66\% on SR,+1.67\% on mAcc and +1.56\% on mIoU ) when $t=4$.
The results suggest that our model performs well with different scales.

 \subsubsection{Comparisons on cross-dataset generalization}
Unlike to previous methods that only {evaluate performance within a dataset}, we provide cross-dataset comparisons in \Cref{tab:cross}, where methods are trained on one dataset and evaluated on another.
(1) We generate free-form procedure planning outputs, treating them as step captions. These captions, along with new textual action step labels, are encoded into vectors using a frozen language encoder. We then retrieve the top $T$ action labels based on the similarity between captions and action labels, where $T$ represents the number of steps in a sequence. We applied a similar encoding approach to step IDs.
(2) Our method outperforms the previous best method \cite{niu2023schema} across different datasets and metrics, likely because PlanLLM generates free-form planning descriptions by incorporating LLMs into video procedure planning.
(3) Both methods perform better on ``train on COIN, test on CrossTask" compared to ``train on CrossTask, test on COIN." This might be due to COIN having more samples than CrossTask, leading to more optimal models when trained on COIN, but testing on COIN presents more challenges than testing on CrossTask.

\subsection{Ablation Studies}

\subsubsection{Effectiveness of proposed components}
In \Cref{tab:component}, we evaluate the effectiveness of our proposed components by establishing a baseline method without Mutual Information Maximization (MIM) or LLM-enhanced planning (1st row, $w/o$ MIM/LLM).
(1) Adding the MIM module (2nd row, $w/o$ LLM) improves performance across various metrics and prediction horizons (+0.96/0.56\% on SR with $t=3/4$), showing its ability to integrate visual cues with commonsense information from textual embeddings, enhancing video procedure planning.
(2) Introducing the LLM component (3rd row, $w/o$ MIM) further boosts performance compared to the baseline (+1.00/1.38\% on SR with $t=3/4$), indicating that the LLM provides valuable planning priors and improves action sequence accuracy.
(3) Combining both MIM and LLM modules results in the greatest improvement (last row, +1.71/3.68\% on SR with $t=3/4$), highlighting their complementarity and collective effectiveness in enhancing video procedure planning.

\subsubsection{Effectiveness of progressive multi-modal training}
In \Cref{tab:progressive}, we validate the effectiveness of progressive multi-modal training.
(1) We {create} a variant with a frozen LLM (1st row), where all LLM parameters are fixed during training. Progressive training (3rd row) outperforms the frozen LLM, indicating that fine-tuning LLM parameters is more effective for video procedure planning.
(2) One-stage training (2nd row) does not achieve optimal results compared to progressive training. Our method's two-stage approach allows the frozen LLM in the first stage to better align embeddings with the input space, and in the second stage, training the LLM optimizes it for downstream tasks. This confirms the effectiveness of progressive training in  planning.

\subsubsection{Different planning generation strategy}
In \Cref{tab:generation}, we compare the effectiveness of different planning generation strategies. We create two variants: one using only the LLM (without knowledge fusion, step decoder, and step refiner) to generate action sequences (``LLM only", 1st row), and another using only step prediction without the LLM and knowledge fusion module (``Step only", 2nd row).
(1) ``Step only" outperforms ``LLM only," likely because LLM's free-form outputs may introduce bias in a closed set, such as inconsistent captions with action labels in the test set.
(2) The best result is achieved by combining information from both branches using the Knowledge Fusion module (3rd row), demonstrating that the two channels complement each other.

\section{Conclusion}
\label{sec:end}
In this paper, we present PlanVLM, a multi-modal model designed to enhance LLMs' planning and visual perception. Our method integrates an LLM with a progressive training strategy to align visual and textual tokens. We also introduce a Mutual Information Maximization module to connect commonsense step descriptions with specific visual states, improving LLMs' reasoning abilities for generating coherent step sequences. Experimental results on three datasets demonstrate the effectiveness of our approach.
\section{Acknowledgements}
This work was supported by the grants from the National Natural Science Foundation of China 62372014.

\bibliography{aaai25}

\end{document}